%% file: main.tex
\documentclass[10pt,twocolumn,letterpaper]{article}

\usepackage[pagenumbers]{cvpr} % To force page numbers, e.g. for an arXiv version

\input{preamble}

\definecolor{cvprblue}{rgb}{0.21,0.49,0.74}
\usepackage[pagebackref,breaklinks,colorlinks,allcolors=cvprblue]{hyperref}

\def\confName{CVPR}
\def\confYear{2027}

\title{ResoLUT: Distilling Residual Networks into Lookup-Table Networks}

\author{
Katarzyna Fojcik$^{1*}$ \quad
Aidas Mikalauskas$^{2,3*}$ \quad
Matthias Kampa$^{2,4}$\\
Renaldas Zioma$^{2}$ \quad
Jogundas Armaitis$^{2}$\\[2mm]
$^{1}$Wroclaw University of Science and Technology \qquad
$^{2}$Transcendent Logic \\
$^{3}$Vilnius University \qquad
$^{4}$Zurich University of Applied Sciences (ZHAW)\\[1mm]
{\tt\small katarzyna.fojcik@pwr.edu.pl} \quad
{\tt\small aidas.mikalauskas@ff.stud.vu.lt \quad kamp@zhaw.ch} \\
{\tt\small rz@TranscendentLogic.com \quad ja@TranscendentLogic.com}\\[1mm]
{\tt\small $^{*}$Equal contribution}
}

\begin{document}
\maketitle
\input{sec/0_abstract}
\input{sec/1_intro}

\input{sec/2_related}
\input{sec/3_method}
\input{sec/4_experiments}
\input{sec/5_discussion}
{
    \small
    \bibliographystyle{ieeenat_fullname}
    \bibliography{main}
}


\end{document}

%% file: preamble.tex
\usepackage{microtype}

\usepackage{multirow}
\usepackage{tikz}
\usetikzlibrary{arrows.meta,positioning,fit,calc,backgrounds}

\newcommand{\method}{ResoLUT\xspace}

%% file: sec/0_abstract.tex
\begin{abstract}
Logic networks built from Boolean gates or lookup tables (LUTs) promise efficient hardware inference, but training them at depth and learning their connectivity remain difficult.
We do not train logic as logic.
Instead, ResoLUT distills a trained residual network (ResNet) layer by layer into LUT trees, preserving its multi-bit residual stream and BatchNorm while extracting connectivity from the teacher rather than learning it.
To our knowledge, ResoLUT is the first convolutional logic network in computer vision to retain a ResNet-style multi-bit residual stream.
Hard LUT decisions are used throughout training and inference, avoiding a discretization gap in the logic.
ResoLUT-20 reaches 95.66\% on CIFAR-10 with twice the trees, and ResoLUT-56 reaches 78.87\% on CIFAR-100 and 53.69\% on ImageNet at 64\,px, the best results of logic and lookup-table networks on CIFAR-10 and CIFAR-100, showing that LUT networks can scale in depth.
We explore and quantify the role of the residuals and the tree budget and shape.
The most optimized model, a ResoLUT-8 with one eighth of the trees and further optimizations, fits on an FPGA and reaches 90.04\% at 189k images per second, the highest throughput among the compared models above 85\% on CIFAR-10.
\end{abstract}

%% file: sec/1_intro.tex
\section{Introduction}
\label{sec:intro}

Digital hardware computes with logic gates and lookup tables (LUTs).
A machine learning model that is \emph{made of} such primitives needs no multipliers, no weight memory traffic and no floating point operations, and it can be unrolled into a single pipelined circuit.
Logic gate networks (LGNs) pursue this idea: Petersen \etal~\cite{petersen2022difflogic} relax each two-input gate to a distribution over the 16 Boolean functions, train with gradient descent and discretize afterwards.
Convolutional logic trees~\cite{petersen2024convlogic}, learnable wiring~\cite{bacellar2024dwn,fojcik2026lilogic} and better relaxations~\cite{yousefi2025gap} have since made LGNs more accurate and cheaper to train.
LUT networks generalize the node from a two-input gate to a $K$-input lookup table~\cite{wang2019lutnet,umuroglu2020logicnets,bacellar2024dwn} (K=2 for LGN).

\begin{figure}[t]
  \centering
   \includegraphics[width=\linewidth]{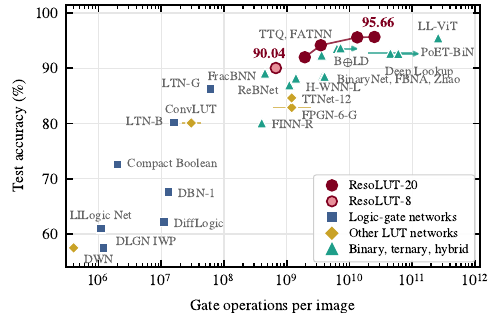}
  \caption{CIFAR-10 test accuracy against gate operations per image: \method{} at several tree budgets (red) and published networks (bars and binary or hybrid networks: estimated gate operations).}
  \label{fig:teaser}
\end{figure}

The accuracy of logic networks, however, has not followed the progress of conventional networks, and the obstacle is depth.
A logic layer maps bits to bits, and every gate that is not a pass-through discards part of its input: the image is represented by progressively fewer independent bits, and the gradient that reaches the early layers, a product of many relaxed gates, vanishes.
In line with this, the discretization gap of the relaxation grows with depth~\cite{yousefi2025gap}, LILogic Net stops improving after two to four layers~\cite{fojcik2026lilogic,fojcik2026towards}, stacked LUT layers lose accuracy rapidly in the pilot runs of BitLogic~\cite{buhrer2026bitlogic}, and even with dedicated initializations the standard gate parametrization loses accuracy with depth, while a reparametrization that avoids this gains little from further depth~\cite{ruttgers2025light}.
The existing remedies keep the residual path at a single bit: residual initializations start every gate as a pass-through so that a deep network is initially an identity~\cite{petersen2024convlogic}, logical skip connections combine the input and the output of a block with a Boolean gate~\cite{kim2024dbn}, and explicit skip paths of fixed pass-through gates do not improve on the initialization~\cite{ruttgers2025light}.
Residual initialization does not guarantee a persistent pass-through, since training may discard it.
None of these remedies enables a block to add a graded correction.
A separate difficulty is connectivity: every gate chooses its inputs from a combinatorial space, and learnable connectomes were found to scale poorly beyond flat networks~\cite{fojcik2026lilogic,fojcik2026towards}.

In conventional networks the corresponding problem was solved by BatchNorm~\cite{ioffe2015bn} and the residual stream~\cite{he2016resnet}.
We ask whether the numerical originals can be kept in a logic network: a multi-bit residual stream and BatchNorm.
Our answer, \method, rethinks a residual network as a network of four-input lookup tables (\cref{fig:method}).
Instead of training logic as logic, it follows the teacher--student idea~\cite{hinton2015distill,romero2015fitnets}: a trained ResNet is converted layer by layer, and each LUT layer learns to reproduce the output of the layer it replaces.
The residual stream runs next to the LUT layers as a low-precision integer per channel and position; each LUT layer ends in a counter whose count, rescaled by a per-channel affine map that absorbs BatchNorm, is added to the stream; the next layer reads the stream through fixed comparators.
Every learned operation remains a lookup table, while the stream keeps a graded value instead of a single bit, so a block can apply a small correction without re-encoding its input.
Connectivity is \emph{extracted} from the teacher's weights, which turns a combinatorial search into a sequence of small regressions, and only truth tables are trained, with a forward pass that is the exact hard network.

We find that:
\begin{itemize}
  \item \method{} reaches 95.66\% on CIFAR-10 and 78.87\% on CIFAR-100, test accuracy, the best among the methods we compare with (\cref{tab:main});
  \item the residual stream matters more the deeper the network: a ResNet-32 without residuals loses 5.0 points in conversion against 3.2 with residuals (CIFAR-100), a teacher with residuals converted into a network without them loses 6.4 points at 20 layers and 15.8 at 32 (CIFAR-100), and a 3-bit stream loses nothing after brief fine-tuning (\cref{sec:depth});
  \item the same stream removes the depth degradation of flat LILogic Net (MNIST, eight layers: 61.6\% to 97.8\%), so the mechanism is not specific to convolutions (\cref{sec:lilogic});
  \item even though the cost in gate operations is high, \method{} lies on a competitive accuracy--gate-operations Pareto front (\cref{fig:teaser,tab:main});
  \item the exported networks run in integers only, and a compact variant processes 189k FPS on one FPGA (\cref{sec:hardware}).
\end{itemize}

\paragraph{Contributions.}
(1)~A LUT convolution with teacher-derived wiring and an integer multi-bit residual stream, trained layer by layer with a hard forward pass.
(2)~Ablations of the residual stream (depth, residuals, per block, bit width) in ResoLUT and in flat LILogic Net, and of tree budget and shape, comparators and wiring.
(3)~The best accuracy among the compared methods on CIFAR-10 and CIFAR-100, with a competitive accuracy--gate-operations Pareto front, and an FPGA implementation.
(4)~An open-source implementation of the full framework (to be released).

%% file: sec/2_related.tex
\section{Related Work}
\label{sec:related}

\paragraph{Differentiable logic gate networks.}
Petersen \etal~\cite{petersen2022difflogic} train randomly wired layers of two-input gates by relaxing each gate to a softmax over the 16 Boolean functions.
Their convolutional extension~\cite{petersen2024convlogic} uses trees of gates in sliding windows, logical-OR pooling and residual initializations (every gate starts as a pass-through, a one-bit, learnable stand-in for a residual connection), and reaches 86.3\% on CIFAR-10 with 61M gates (LogicTreeNet-G); it remains the strongest logic network trained from scratch.
Differentiable weightless networks~\cite{bacellar2024dwn} learn LUT contents and address mappings; LILogic~\cite{fojcik2026lilogic} makes the connectome itself learnable with Top-K candidates and a basis projection that speeds up training, but is limited to flat networks of at most a few hundred thousand gates: its dense variants peak at two layers, and penalizing pass-through gates does not make a four-layer network better than a single layer~\cite{fojcik2026towards}.
Yousefi \etal~\cite{yousefi2025gap} show that Gumbel noise with a straight-through estimator closes the discretization gap and speeds up training, but report that the gap grows with depth for the standard relaxation; R\"uttgers \etal~\cite{ruttgers2025light} shrink the gate parametrization yet still find that depth harms accuracy; LogicIR~\cite{lee2026logicir} uses a UNet of gates for image restoration.
All of these train logic from scratch with a soft forward pass, either fix their wiring at random or learn it alongside the gates, and carry a single bit along any residual path they have. We instead start from a trained network: the forward pass stays hard, the wiring comes from the teacher, and the residual stream keeps multiple bits.

\paragraph{LUT networks.}
LUT networks replace the two-input gate of an LGN with a $K$-input lookup table, of which the gate is the $K{=}2$ case.
LUTNet~\cite{wang2019lutnet} replaces binary multiply-accumulates by learned $K$-input LUTs; LogicNets~\cite{umuroglu2020logicnets}, PolyLUT~\cite{andronic2023polylut} and NeuraLUT~\cite{andronic2024neuralut} constrain neuron fan-in so that each neuron becomes a truth table; TTNet~\cite{ijcai2024p2} converts a CNN with Heaviside activations into truth tables and then into logic; ConvLUT~\cite{andronic2026convlut} builds fully unrolled convolutional networks from hierarchical LUT trees; Deep Lookup Network~\cite{guo2025deeplookup} replaces the multiplications of a numerical CNN by learned table lookups and keeps its additions.
LL-ViT~\cite{nag2025llvit} replaces the feed-forward blocks of a transformer by LUT layers that sit next to its real-valued residual stream, with attention left numerical.
BitLogic~\cite{buhrer2026bitlogic} compares encoders, connectivity rules and fan-ins of 2, 4 and 6 for two-layer feed-forward networks, finds 4 to be the accuracy--cost knee, as we do, and names depth and residual LUT networks as open.
We differ in what is converted and how: we replace \emph{every} weighted layer of a deep convolutional ResNet, and the 4-input LUT is the only learned object.
The multi-bit residual stream and BatchNorm are retained in integer form rather than binarized.

BinaryNet~\cite{courbariaux2016bnn}, XNOR-Net~\cite{rastegari2016xnor}, Bi-Real Net~\cite{liu2018bireal}, ReActNet~\cite{liu2020reactnet}, ReCU~\cite{xu2021recu}, AdaBin~\cite{tu2022adabin} and BNext~\cite{guo2022bnext} constrain weights and activations to $\pm1$ and recover most of the accuracy of real-valued networks; ternary networks~\cite{zhu2017ttq} sit between the two, and FINN~\cite{umuroglu2017finn} compiles binary networks to FPGAs.
These networks are the natural accuracy reference, and in \cref{sec:pareto} we compare gate operations with them; they keep the multiply-accumulate as the unit of work, whereas our layers contain none.

\paragraph{Distillation and layer-wise conversion.}
Knowledge distillation~\cite{hinton2015distill} and intermediate-feature hints~\cite{romero2015fitnets} transfer a trained network into a differently structured one.
Distillation has been applied to differentiable logic networks on tabular data~\cite{suh2026kd}, where an MLP teacher guides a logic student that is trained as usual.
We use both signals, logits and per-layer output matching, but convert a convolutional teacher one layer at a time: each new layer is trained against the numerical layer it replaces, on the inputs it will receive from the converted prefix, with straight-through gradients~\cite{bengio2013ste} through the hard lookups.

%% file: sec/3_method.tex
\section{Method}
\label{sec:method}

\begin{figure*}[t]
\centering
\begin{subfigure}{\textwidth}
  \centering
  \includegraphics[width=\linewidth,trim=2bp 2bp 2bp 2bp,clip]{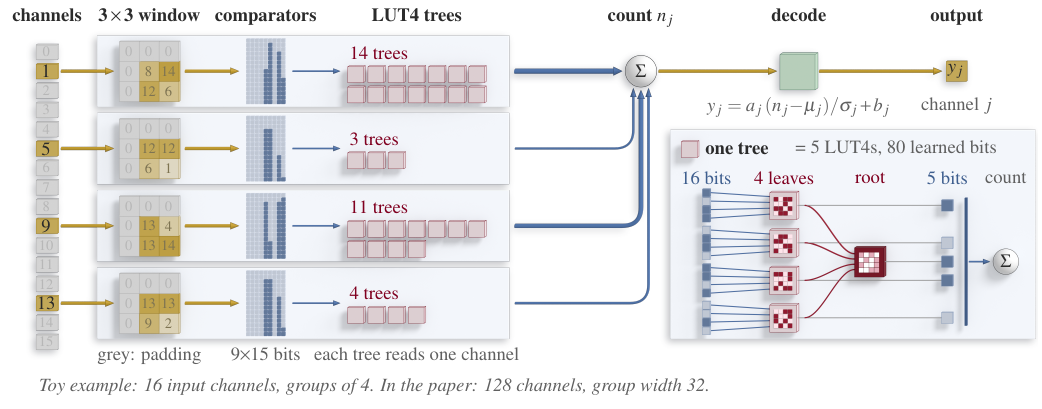}
  \caption{One LUT convolution, one output channel. The width of a lane's line into the count follows its number of trees.}
  \label{fig:method_a}
\end{subfigure}\\[4pt]
\begin{subfigure}{\textwidth}
  \centering
  \includegraphics[width=\linewidth,trim=2bp 2bp 2bp 2bp,clip]{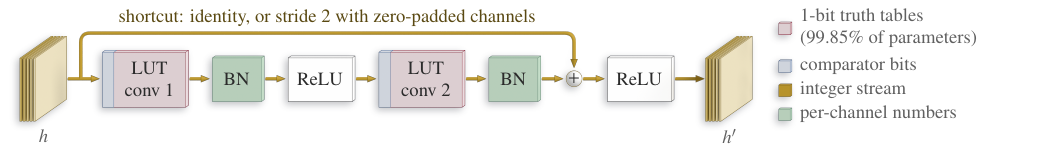}
  \caption{Residual block with the integer stream.}
  \label{fig:method_b}
\end{subfigure}\\[4pt]
\begin{subfigure}{\textwidth}
  \centering
  \includegraphics[width=\linewidth,trim=2bp 2bp 2bp 2bp,clip]{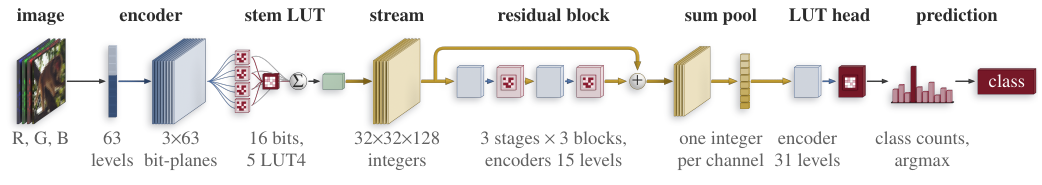}
  \caption{A complete network from image to class (CIFAR-100 ResoLUT-20).}
  \label{fig:method_c}
\end{subfigure}
\caption{The LUT convolution (a), its place in a residual block (b), and a complete network (c). Comparators turn the integer residual stream into thermometer bits; each tree reads 16 of them from one input channel's window and its five LUT outputs are counted; channels mix only in this count. A per-channel affine map decodes the count back into the convolution's output, and BatchNorm, the shortcut addition and ReLU are the teacher's own, in fixed point. Pink: LUT4 truth tables, the only learned bits; slate: comparator bits; gold: integers; green: per-channel numbers.}
\label{fig:method}
\end{figure*}

ResoLUT takes a trained ResNet, the \emph{teacher}, and returns a network of LUTs.
This section sets out the teachers and the layers we replace (\cref{sec:setting}) and the residual stream that connects them (\cref{sec:stream}), then covers the LUT convolution (\cref{sec:lutconv}), its wiring (\cref{sec:routing}) and how each layer is trained (\cref{sec:training}).
\Cref{fig:method} gives an overview; \cref{tab:models} lists the teachers and converted models, and the supplement gives the hyperparameters of every model.

\subsection{Teachers and replaced layers}
\label{sec:setting}
\paragraph{Teacher architecture.}
The teachers are ordinary floating-point ResNets~\cite{he2016resnet}, trained once and frozen: a $3{\times}3$ stem convolution (3 to $w_1$ channels), three stages of $n$ residual blocks each ($w_1$, $w_2$ and $w_3$ channels), global average pooling and a linear classifier (the \emph{head}).
The network has $6n+2$ weighted layers, so $n=1,3,5,9$ gives ResNet-8, 20, 32 and 56.
Every block is the original \emph{basic block}: two $3{\times}3$ convolutions with BatchNorm after each, a ReLU after the first, and one residual connection that bypasses both and is added before the final ReLU; only the first block of stages 2 and 3 differs (stride 2, doubled channels, and a parameter-free option-A shortcut that subsamples and zero-pads).

\paragraph{Widths and grouped convolutions.}
The teachers are 4 to 12 times wider than the original $16/32/64$ channels. 
% $w=128/256/512$ for the CIFAR-10 models (and a narrow ResNet-8 with $64/128/256$), the CIFAR-100 ResNet-8, 32 and 56 and the ImageNet-64 models (stride-2 stem), and $192/384/768$ for the CIFAR-100 ResNet-20.
% The ResNet-8 thus runs at $32{\times}32$ with 128 channels, then $16{\times}16$ with 256 and $8{\times}8$ with 512 (\cref{fig:method}c).
Block convolutions are \emph{grouped}: a fixed 0/1 mask lets each output read only 32 of its input channels (interleaved groups in the first convolution of a block, contiguous groups in the second, so that together they mix all channels); the low fan-in keeps what a LUT layer must reproduce simple.

\paragraph{What is replaced.}
Every weighted layer is replaced by a LUT layer: the stem, both convolutions of every block, and the head (a linear map on the pooled features, \ie a $1{\times}1$ convolution).
A teacher ResNet-$L$ becomes a network of $L$ LUT layers; the residual connections stay where they are.

\paragraph{Teacher training.}
CIFAR teachers are trained for 600 epochs with Nesterov SGD (learning rate 0.1, cosine decay, weight decay $5{\times}10^{-4}$, batch 128), flips and 4-pixel translations, mixup~\cite{zhang2018mixup}, and distillation~\cite{hinton2015distill} (weight 0.9, temperature 2) from an airbench-style network~\cite{jordan2024airbench} for the first ResNet-8 and from an ensemble of four WRN-28-10~\cite{zagoruyko2016wrn} trained with SAM~\cite{foret2021sam} for all later teachers; ImageNet-64 teachers are distilled from a ResNet-50 for 64 epochs.
Gaussian noise of 0.3 channel standard deviations after each block's BatchNorm makes the teacher tolerate the approximation error of lookup tables.
The converted network follows its teacher, so most accuracy gains came from better teachers (supplement).

\subsection{The residual stream}
\label{sec:stream}
Between LUT layers the activations are not bits but a \emph{residual stream}: a low-precision integer per channel and spatial position, an explicit component of the architecture.
A LUT layer does not write its output to the stream directly: each output channel ends in a counter that sums the outputs of its lookup tables (\cref{sec:lutconv}), and a per-channel scale and offset, into which the teacher's BatchNorm is folded, turn the count into the block's update, which is added to the stream.
The next layer reads the stream through a per-channel thermometer code of fixed thresholds.
Every learned operation remains a lookup table, while the stream keeps a graded value instead of a single bit, so a block can apply a small correction without re-encoding its input.
Resolution-changing shortcuts are parameter-free (option~A~\cite{he2016resnet}), so the residual path has no learned parameters.
The residual additions, the per-channel decode and the final pooling are the only arithmetic; all are integer operations, and the exported network runs in integer arithmetic only.
The stream carries $f=8$ fractional bits in all exported models; its width matters for hardware, since every bit is a wire and an adder bit (\cref{sec:depth}).

\subsection{The LUT convolution}
\label{sec:lutconv}
A convolution computes, for every output channel and output position, a sum over the input channels of a $3{\times}3$ filter applied to a window of the input.
A LUT convolution keeps this structure (sliding window, stride, zero padding, one set of tables per output channel shared by all positions) and replaces the multiply-and-add by three steps (\cref{fig:method}a): comparators, trees and a vote count.
The head is the special case of a $1{\times}1$ window on the pooled features. On ImageNet-64 the head instead holds one small table per class and pooled channel: 15 comparators give each channel a 4-bit code, a table entry is the numerical head's weight times the mean feature of that code, rounded to an integer, and a class score is the sum of its entries; it is fitted without training and uses the same number of LUT4s as the tree head.
\paragraph{Comparators.}
Each input value is compared with $K$ cutoffs, and bit $k$ is 1 if the value exceeds cutoff $k$.
The bits therefore form a \emph{thermometer code}: ones up to the value's level, zeros above it.
$K$ bits distinguish only $K{+}1$ levels, so they carry $\log_2(K{+}1)$ bits of information ($K=15$: 16 levels, 4 bits).
A $3{\times}3$ window thus offers $9K$ bits per input channel.
The cutoffs are uniform grids chosen per layer type ($K=15$ in the blocks, 63 in the stem, 15 or 31 in the head; see the supplement) and are not learned.

\paragraph{Trees.}
Every output channel owns $V$ trees.
A tree reads 16 of the $9{\times}K$ bits of a \emph{single} input channel (any window position and level, each optionally inverted) through four \emph{leaf} LUT4s that read four bits each and a \emph{root} LUT4 that combines the four leaf outputs; it outputs five bits.
Which bits a tree reads is its \emph{wiring}, fixed before training (\cref{sec:routing}).
A LUT4 stores one output bit for each of its $2^4=16$ input patterns, so a tree of five LUT4s stores $5{\times}16=80$ table bits; these are the only learned bits (28.97M of the 29.02M parameters of the CIFAR-10 ResoLUT-8).
% This shape (4 inputs per LUT, 16 per tree) is shared by all layers and models: the LUT4 is a native FPGA primitive and BitLogic~\cite{buhrer2026bitlogic} also finds 4 to be the accuracy--cost knee among fan-ins 2, 4 and 6; other shapes are in \cref{sec:ablations}.
All layers and models share this shape: the LUT4 is a native FPGA primitive and BitLogic~\cite{buhrer2026bitlogic} also finds 4 inputs to be the accuracy--cost knee among fan-ins 2, 4 and 6 (see \cref{sec:ablations}).

\paragraph{Vote count and decode.}
All trees of output channel $j$ vote with their five output bits, and the layer's integer output $n_j$ is the number of ones, between 0 and $5V$.
The count plays the role of the sum over input channels: each tree is a nonlinear filter of one channel's window, the number of trees given to a channel mimics the importance of its filter weights (\cref{sec:routing}), and channels mix only in the count.
A per-channel affine map $y_j=a_j\,(n_j-\mu_j)/\sigma_j+b_j$ turns the count back into the replaced convolution's output ($\mu_j,\sigma_j$: mean and standard deviation of the count before training; $a_j,b_j$: a learned scale and offset).
$V$ is a per-layer budget (supplement): by default 64 in the stem, 512 per class in the head and, in block convolutions, 8 per input channel that an output reads (\ie 256, since a grouped output reads 32 channels); the CIFAR-10 ResoLUT-8 uses hand-set values (\cref{fig:method}c).
For integer-only inference, every comparator threshold is read off the trained model exactly, and the decoder, the following BatchNorm and the shortcut collapse into one integer multiply-add and a shift per channel; the only approximation is rounding the stream to $f$ fractional bits, which changes accuracy by 0.0 to 0.3 points (supplement).

\subsection{Connectivity from the teacher}
\label{sec:routing}
Learning which of the $9{\times}K$ bits each tree should read is the expensive part of a logic network.
We make this choice once, before training, from the teacher and 512 training images.
For every output channel $j$ and input channel $c$ we take $q_{jc}$, the part of the teacher's output that comes from this input channel (its filter applied to its input; the output is the sum of these parts), computed on the activations the layer will receive through the converted prefix at 16 positions per image.
First, the trees of an output are split over its input channels in proportion to the standard deviation of $q_{jc}$ (rounded by largest remainder), so channels that matter get more trees and channels with zero teacher weight get none.
Second, a tree draws its 16 bits without replacement from the $9{\times}K$ bits of its channel, with probability proportional to their absolute correlation with $q_{jc}$, and inverts every bit whose correlation is negative.
Thermometer bits of one value are nested, so a LUT that reads two of them sees at most 12 of its 16 input patterns; the 16 bits are therefore dealt so that each leaf LUT reads four \emph{different} window positions.
Every table starts as a pass-through of its first pin, as in residual initializations~\cite{petersen2024convlogic}, so a tree is initially its first routed bit and the decoder can be fit to the teacher's output by least squares before any training.

\subsection{Layer-wise conversion}
\label{sec:training}
Layers are replaced in forward order, stem first.
At stage $\ell$ the layers before $\ell$ are frozen LUT convolutions, layer $\ell$ is new, and the layers after it are still the teacher's numerical layers (a scaffold).
The new layer is calibrated on the activations it receives \emph{through the converted prefix}, so it learns to correct earlier errors, and is trained with
\begin{equation}
\mathcal{L}_\ell=\tfrac12\,\mathrm{CE}(z,y)+\tfrac12\,T^{2}\,\mathrm{KL}\!\bigl(\sigma(z^{\mathrm T}/T)\,\big\|\,\sigma(z/T)\bigr)+\tfrac12\,\mathrm{MSE}_\ell,
\end{equation}
where $z,z^{\mathrm T}$ are the logits of the partly converted network and of the teacher ($T{=}2$) and $\mathrm{MSE}_\ell$ compares the layer's output with that of the numerical convolution it replaces on the same input, per channel normalized by its calibration standard deviation; without this term accuracy falls by more than a point (90.68 against about 92\%, CIFAR-10 ResoLUT-8).
Each LUT entry has a real score $s\in[-1,1]$ and the bit is $[s\ge0]$.
The forward pass is the hard lookup, identical to the exported network, so \emph{there is no discretization gap}; the backward pass sends each output's gradient to the score of the entry it read and to every pin as the finite difference of the table across that pin~\cite{bengio2013ste}.
We use AdamW~\cite{loshchilov2019adamw} (learning rate $0.02$, cosine decay to 1\%, 50 warm-up updates, no weight decay), batch 32 for the CIFAR-10 ResoLUT-8 and 128 otherwise, and 6, 8 and 10 epochs per layer in the three stages (6 for the stem and the head; one on ImageNet).
The scaffold and the BatchNorm affines adapt at $3{\times}10^{-4}$ until their layer is replaced; an optional joint phase trains all tables together and helps only the smallest models (\cref{sec:ablations}).

%% file: sec/4_experiments.tex
\section{Experiments}
\label{sec:experiments}

\subsection{Setup}
CIFAR-10 and CIFAR-100~\cite{krizhevsky2009cifar} are split into 45,000 training, 5,000 validation and 10,000 test images; ImageNet~\cite{russakovsky2015ilsvrc} (1,000 classes, 64$\times$64 pixels~\cite{chrabaszcz2017downsampled}) into 1,256,167 training, 25,000 validation and 50,000 test images (the official validation set).
We report test accuracies as the mean $\pm$ sample standard deviation over three seeds where a deviation is given, and a single run otherwise.
The supplementary material gives the exact numbers behind all graphs and the results of further settings.
\emph{LUT4} counts the four-input lookup tables of the trees (logical count).
\emph{Gate operations} are two-input gate operations per image, the measure of LogicTreeNet~\cite{petersen2024convlogic}: each layer is synthesized with Yosys~\cite{wolf2013yosys} into LUT4 logic (trees, counters, comparators, decoder), times the kernel placements, at three gates per LUT4; binary/ternary networks without a published count use 8 gates per binary multiply-accumulate.

\subsection{Baseline results}
\label{sec:main}

\begin{table*}[t]
\centering
\footnotesize
\setlength{\tabcolsep}{4pt}

\begin{tabular}{@{}lllllrrrr@{}}
\toprule
Dataset & Model & Teacher widths & Teacher source & Training time & Teacher acc. & ResoLUT acc. & Gap & LUT4 \\
\midrule
CIFAR-10  & ResoLUT-8  & 128/256/512 & airbench-style net & 0.7\,h & 93.42 & 92.32 $\pm$ 0.14 & $-$1.10 & 1.81M \\
          & ResoLUT-20 & 128/256/512 & 4$\times$WRN-28-10 & 4.6\,h & 96.81 & 95.59 $\pm$ 0.06 & $-$1.22 & 6.39M \\
          & ResoLUT-32 & 128/256/512 & 4$\times$WRN-28-10 & 10.4\,h (2 GPUs) & 96.98 & 95.01 $\pm$ 0.12 & $-$1.97 & 10.38M \\
          & ResoLUT-56 & 128/256/512 & 4$\times$WRN-28-10 & 31.0\,h (2 GPUs) & 97.10 & 95.11 & $-$1.99 & 17.68M \\
CIFAR-100 & ResoLUT-8  & 128/256/512 & 4$\times$WRN-28-10 & 0.8\,h & 73.26 & 71.21 $\pm$ 0.14 & $-$2.05 & 2.09M \\
          & ResoLUT-20 & 192/384/768 & 4$\times$WRN-28-10 & 6.8\,h & 81.20 & 78.16 $\pm$ 0.12 & $-$3.04 & 10.00M \\
          & ResoLUT-32 & 128/256/512 & 4$\times$WRN-28-10 & 16.2\,h & 81.67 & 78.21 $\pm$ 0.37 & $-$3.46 & 11.14M \\
          & ResoLUT-56 & 128/256/512 & 4$\times$WRN-28-10 & 57.3\,h & 83.44 & 78.87 $\pm$ 0.28 & $-$4.57 & 18.60M \\
ImageNet-64 & ResoLUT-20 & 128/256/512 & ResNet-50 & 15.8\,h (3 GPUs) & 50.87 & 48.35 & $-$2.52 & 9.48M \\
          & ResoLUT-32 & 128/256/512 & ResNet-50 & 41.4\,h & 54.58 & 51.08 $\pm$ 0.08 & $-$3.50 & 14.07M \\
          & ResoLUT-56 & 128/256/512 & ResNet-50 & 116.5\,h & 58.30 & 53.69 $\pm$ 0.10 & $-$4.61 & 23.24M \\
\bottomrule
\end{tabular}
\caption{Best teacher source and teacher width per ResoLUT depth, converted at the full tree budget. Accuracy (\%) is test accuracy (for ImageNet-64, the 50,000 official validation images): mean $\pm$ std over 3 seeds, or a single run where no deviation is given. Gap = ResoLUT $-$ teacher. Training time is the layer-wise conversion of one run, as a range over seeds, on one RTX 5090 unless noted.}
\label{tab:models}
\end{table*}
\Cref{tab:models} keeps, for every ResoLUT depth, the results for the best teacher source and teacher width we found; all results are at the full tree budget, which \cref{fig:budget} scales from 1/16 to 2$\times$.
The conversion loses accuracy against the teacher (Gap column), and the loss grows with depth on all three datasets: accuracy peaks at 20 layers on CIFAR-10 while on CIFAR-100 and ImageNet it keeps rising up to 56 (\cref{sec:depth}).

\subsection{Residual stream: depth, dependence and width}
\label{sec:depth}
\begin{figure}[t]
\centering
\includegraphics[width=\linewidth]{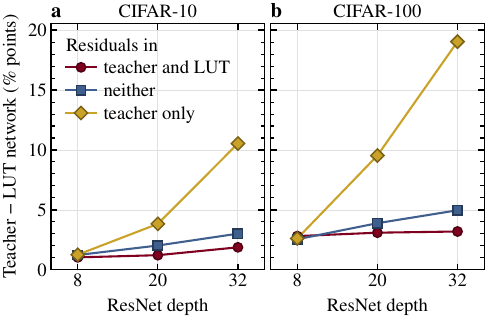}
\caption{Conversion loss (teacher minus LUT network) against depth on CIFAR-10 (a) and CIFAR-100 (b), with residuals in the teacher and ResoLUT, in neither, or in the teacher only.}
\label{fig:skips}
\end{figure}

\begin{figure}[tb]
\centering
\includegraphics[width=\linewidth]{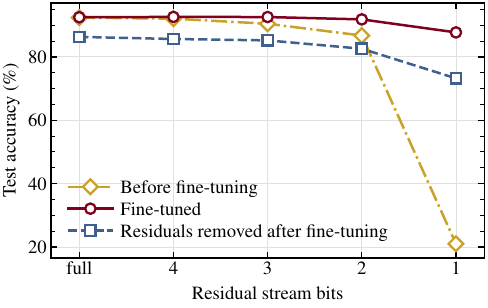}
\caption{Residual stream width, CIFAR-10 ResoLUT-8 (one run): stream rounded to the given number of bits, before and after a five-epoch fine-tune, and with the residuals removed after fine-tuning.}
\label{fig:bits}
\end{figure}

\paragraph{Depth.}
At the smallest depth residuals are not essential: a teacher trained without them converts as well (\cref{fig:skips}b; CIFAR-100 loss 2.6 against 2.8 points for the ResoLUT-8, without against with residuals).
With depth the difference grows: 3.9 against 3.1 points for the ResoLUT-20 and 5.0 against 3.2 for the ResoLUT-32 although the teachers are about as accurate (80.94 vs 81.67\% for the ResoLUT-32): each approximated layer adds error, which accumulates when no stream propagates the signal.

\paragraph{Dependence on the residuals.}
Networks trained with residuals depend on them as they get deeper: converting a teacher with residuals into a LUT network without them is about as good for the ResoLUT-8 but loses 6.4 points at 20 layers and 15.8 at 32 (CIFAR-100; \cref{fig:skips}, teacher only).

\paragraph{Width of the stream.}
We tested how the width of the stream influences accuracy on one example, the CIFAR-10 ResoLUT-8 (one run).
Five epochs of fine-tuning with the rounded stream recover 3 and 4 bits fully (92.66 and 92.65\% test, 92.57\% at full precision) and leave 0.83 points at 2 bits and 4.85 at 1 bit (one seed, \cref{fig:bits}).
How far the stream can be narrowed is probably very specific to the architecture and a next step worth optimizing.
For now we narrow it only for the ResoLUT-8 with one eighth of the trees; all other models keep full stream precision.

\subsection{Residual streams in flat LILogic Net}
\label{sec:lilogic}
% To test whether the stream helps logic networks in general, we add residuals to the fully connected LILogic Net~\cite{fojcik2026lilogic} in three ways: gate inputs may also pick from two layers back (\emph{wire}, one bit); the output of a layer is XORed with its input (\emph{XOR}, one bit); and the counting stream of ResoLUT (\emph{count}: gate outputs are counted, added to a running integer stream and read through comparators).
% We show the two settings of the LILogic Net paper (\cref{fig:lilogic}).
% On MNIST (dense, 4K gates per layer) all three fix the depth problem: at eight layers the plain network collapses to 61.6\% while the residuals recover 92.5, 96.2 and 97.8\%.
% On CIFAR-10 (dense, 8K gates) the residuals keep 46 to 50\% at six layers where the plain network falls to chance, still below the single-layer 53.7\%: the stream stops the degradation but does not make deeper flat networks better.
% The other settings, where the plain network trains better and the residuals help less, and the analysis of why the plain network falls with depth are in the supplement (\cref{sec:suppl_lilogic}).

\begin{figure}[tb]
\centering
\includegraphics[width=0.85\linewidth]{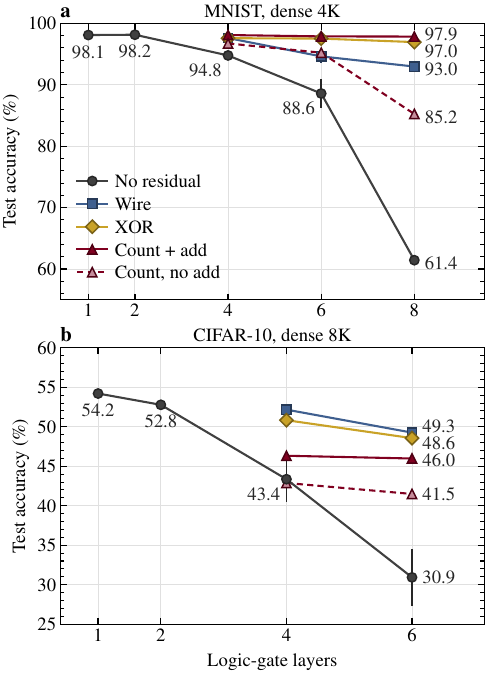}
\caption{Residuals in the fully connected LILogic Net, settings of its paper: binarized test accuracy against the number of logic layers on a) MNIST and b) CIFAR-10. }
\label{fig:lilogic}
\end{figure}

To test whether the stream helps logic networks in general, we add residuals to the fully connected LILogic Net~\cite{fojcik2026lilogic} in three ways: gate inputs may also pick from two layers back (\emph{wire}, one bit), a layer's output is XORed with its input (\emph{XOR}, one bit), or the counting stream of ResoLUT is used (\emph{count}: gate outputs are counted, added to a running integer stream and read through comparators).
On MNIST (dense, 4K gates per layer; \cref{fig:lilogic}) all three fix the depth problem: at eight layers the plain network collapses to 61.6\% while the residuals recover 92.5, 96.2 and 97.8\%.
On CIFAR-10 (dense, 8K gates) they keep 46 to 50\% at six layers where the plain network falls to chance, but stay below the single-layer 53.7\%: the stream stops the degradation without making deeper flat networks better.

\begin{figure}[t]
\centering
\includegraphics[width=0.9\linewidth]{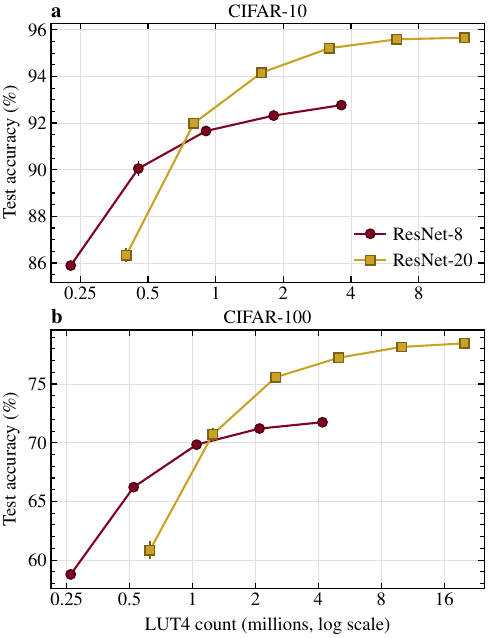}
\caption{Tree budget: test accuracy against the LUT4 count (log scale) on CIFAR-10 (a) and CIFAR-100 (b), with every layer's trees scaled from 1/16 to 2$\times$; mean over three seeds, bars: std.}
\label{fig:budget}
\end{figure}

\paragraph{Budget.}
Accuracy falls gradually with the number of trees (\cref{fig:budget}): on CIFAR-10 it flattens at the full budget (halving costs 0.66 points for the ResoLUT-8 and 0.38 for the ResoLUT-20, doubling gains 0.45 and 0.07), on CIFAR-100 more trees keep helping (1.37 and 0.92 points lost by halving, 0.54 and 0.31 gained by doubling).
An equal LUT4 count does not mean an equal accuracy: with 1.6M LUT4 the ResoLUT-20 reaches 94.16\% on CIFAR-10 against 92.32\% for the ResoLUT-8 with 1.8M, and on CIFAR-100 it comes within 0.5 points of the ResoLUT-8 with 40\% fewer LUT4s (70.71\% with 1.25M against 71.21\% with 2.09M).
Deeper networks pay off only on CIFAR-100, where the ResoLUT-32 and 56 reach 78.21\% and 78.87\% against 78.16\% for the ResoLUT-20 at the full budget, with 11\% and 86\% more LUT4s; on CIFAR-10 the ResoLUT-32 is worse than the ResoLUT-20 at both budgets we ran.
We therefore chose the representative models of \cref{tab:main} from this study: the ResoLUT-8 and 20 at several budgets and, on CIFAR-100, the deeper networks at the full budget.

\subsection{Tree budget and shape}
\label{sec:ablations}

\paragraph{Tree shape and routing.}
In an initial check on the CIFAR-10 ResoLUT-8 we varied the tree shape (eight instead of sixteen inputs, four leaf LUT4s without the root, counting only the root's output) and the routing (random instead of teacher-derived).
All baseline and chosen models keep the default of 16 inputs, four leaf LUT4s plus a root; the other variants were only checked as directions for future optimization.
Counting only the root saves no LUT4s and costs 1.6 points; dropping the root saves 20\% of the LUT4s for only 0.3 points, so it may be a way of optimization; random routing loses 15.6 points.

\subsection{Comparison with the state of the art}
\label{sec:sota}
\begin{table}[tb]
\centering
\footnotesize
\setlength{\tabcolsep}{1.3pt}
\begin{tabular}{@{}llr@{\hspace{5pt}}rr@{}}
\toprule
Method & Type & \shortstack{Acc.\\(\%)} & \shortstack{Size\\(as printed)} & \shortstack{Gate\\operation} \\
\midrule
\multicolumn{5}{@{}l}{\emph{CIFAR-10}} \\
DLGN IWP~\cite{ruttgers2025light} & LGN & 57.47 & 512K LUT2 & $\sim$1.2M \\
DWN~\cite{bacellar2024dwn} & LUT & 57.51 & 48K LUT2 & $\sim$0.4M \\
%DWN~\cite{bacellar2024dwn} & LUT & 57.51 & 45.7K LUT4 & $\sim$0.9-1.3M \\
LILogic Net~\cite{fojcik2026lilogic} & LGN & 60.98 & 256K LUT2 & $\sim$1.1M \\
DiffLogic Net~\cite{petersen2022difflogic} & LGN & 62.14 & 5.12M LUT2 & $\sim$11M \\
DBN-1~\cite{kim2024dbn} & Bool. & 67.60 & 2.47M LUT2 & $\sim$13M \\
Compact Boolean~\cite{wang2026learning} & Bool. & 72.56 & 1.08M LUT2 & $\sim$2M \\
ConvLUT~\cite{andronic2026convlut} & LUT & 80.08 & 2.13M LUT6 & $\sim$21-43M \\
FINN-R CNV-6~\cite{blott2018finnr} & BNN & 80.10  & 1.4Mbit & $\sim$0.4G \\
LogicTreeNet-B~\cite{petersen2024convlogic} & LGN & 80.17 & 16.0M LUT2 & 16.0M \\
% TM Composites~\todo{cite} & Tsetlin & 82.8 & -- & -- \\
FPGN-6-G~\cite{liang2026fpgn} & LUT & 82.9 & 2.46M LUT6 & $\sim$0.6..2.4G \\
TTNet-12~\cite{ijcai2024p2} & hybrid & 84.63 & $\sim$7K LUT12 & 1.2G \\
LogicTreeNet-G~\cite{petersen2024convlogic} & LGN & 86.29 & 61.0M LUT2 & 61.0M \\
ReBNet, 3 residuals~\cite{ghasemzadeh2018rebnet} & BNN & 86.98 & $\sim$1.54Mbit & $\sim$1.1G\\
H-WNN-L~\cite{jadhao2025hybrid} & hybrid & 88.23 & 20.72K LUT6 & $\sim$1.4G \\
Zhao \etal~\cite{zhao2017accelerating} & BNN & 88.54 & 13.4Mbit & $\sim$4.0G \\
BinaryNet~\cite{courbariaux2016bnn} & BNN & 88.60 & $\sim$13.4Mbit & $\sim$3.7G \\
FBNA CNV~\cite{guo2018fbna} & BNN & 88.61 & $\sim$13.4Mbit & $\sim$3.9G \\
FracBNN~\cite{zhang2021fracbnn} & BNN & 89.10 & 0.27Mbit & $\sim$0.45G \\
B$\oplus$LD~\cite{nguyen2024bold} & Bool. & 92.37 & $\gtrsim$30Mbit & $\sim$3.6G \\
PoET-BiN~\cite{chidambaram2020poetbin} & clf. & 92.64 & 9.65KLUT+294Mbit & $\gtrsim$60G \\
Deep Lookup Net~\cite{guo2025deeplookup} & lookup & 92.68 & 17 (33$\times$33) tables & $\sim$20-100G \\
TTQ ResNet-56~\cite{zhu2017ttq} & hybrid & 93.56 & $\sim$0.85Mtern & $\gtrsim$6.3G \\
FATNN VGG-Small~\cite{chen2021fatnn} & TNN & 93.7 & $\sim$4.6Mtern & $\sim$7.2G \\
LL-ViT~\cite{nag2025llvit} & hybrid & 95.5 & 589K LUT & $\sim$260G \\
\cmidrule{1-5}
ResoLUT-8, 1/8 trees + opt. & LUT & 90.04 & 0.22M LUT4 & 0.67G \\
ResoLUT-20, 1/8 trees & LUT & 91.99 & 0.80M LUT4 & 1.94G \\
ResoLUT-20, 1/4 trees & LUT & 94.16 & 1.60M LUT4 & 3.49G \\
ResoLUT-20 & LUT & 95.59 & 6.39M LUT4 & 13.3G \\
ResoLUT-20, 2$\times$ trees & LUT & \textbf{95.66} & 12.79M LUT4 & 25G \\
\midrule
\midrule
\multicolumn{5}{@{}l}{\emph{CIFAR-100}} \\
DLGN IWP~\cite{ruttgers2025light} & LGN & 29.5 & 1.54M LUT2 & $\sim$2.3M \\
% Yousefi \etal~\cite{yousefi2025gap} & LGN & 28. & -- & $\sim$4.5M \\
DBN-1~\cite{kim2024dbn}  & Bool. & 39.46 & 19.80M LUT2 & $\sim$116M \\
% TM Composites~\todo{cite} & Tsetlin & 52.2 & -- & -- \\
B$\oplus$LD~\cite{nguyen2024bold} & Bool. & 68.43 & $\gtrsim$30Mbit   & $\sim$3.6G \\
Deep Lookup Net~\cite{guo2025deeplookup} & lookup & 68.96 & 17 (33$\times$33 tables) & $\sim$20-100G \\
BNext-18~\cite{guo2022bnext} & BNN & 72.18 & $\sim$11.2Mbit & $\sim$3.3G \\
LL-ViT~\cite{nag2025llvit} & hybrid & 78.8 & 589K LUT & $\sim$260G \\
\cmidrule{1-5}
ResoLUT-20, 1/8 trees& LUT & 70.71 & 1.25M LUT4 & 3.24G \\
ResoLUT-8 & LUT & 71.21 & 2.09M LUT4 & 3.50G \\
ResoLUT-20, 1/2 trees& LUT & 77.24 & 5.00M LUT4 & 11.1G \\
ResoLUT-20 & LUT & 78.16 & 10.00M LUT4 & 21.3G \\
ResoLUT-32 & LUT & 78.21 & 11.14M LUT4 & 24.3G \\
ResoLUT-56 & LUT & \textbf{78.87} & 18.60M LUT4 & 39.1G \\
\bottomrule
\end{tabular}
\caption{Comparison with the state of the art (test accuracy). Ours, lower block: mean over seeds at tree budgets 1/8, 1/4, 1/2, 2$\times$. Others: published; $\sim$: estimate. Size as printed (ours: LUT4 count); gate operations: two-input gates per image. Hybrid, lookup, Bool.: numerical parts remain; clf.: classifier only.}

\label{tab:main}        
\end{table}

\Cref{tab:main}: networks made only of gates and LUTs, trained from scratch, are clearly less accurate at every size; only networks that keep numerical parts come close, and the ResoLUT-20 (CIFAR-10) and ResoLUT-56 (CIFAR-100) conversions are above all compared methods.

% In \cref{tab:main} the ResNet-20 conversion reaches 95.59\% on CIFAR-10, 9.3 points above LogicTreeNet-G, and 78.16\% on CIFAR-100 against about 31\% for logic networks trained from scratch; on ImageNet at 64\,px the 20-, 32- and 56-layer networks reach 48.35, 51.03 and 53.78\%, where the only Boolean-network ImageNet number we know is 41.6\% for TTNet~\cite{benamira2022ttnet} \todo{check its protocol}.
% The strongest published lookup-table results keep numerical parts: LL-ViT~\cite{nag2025llvit} replaces the MLPs of a vision transformer at 224\,px, and Deep Lookup Network~\cite{guo2025deeplookup} turns the multiplications of a ResNet-20 into lookups; ours is 2.9 points above the latter on CIFAR-10 and, with the ResNet-20 at twice the trees on CIFAR-10 and the ResNet-56 on CIFAR-100, above LL-ViT (95.66 and 78.87\% against 95.5 and 78.8\%), converting every weighted layer.

\subsubsection{Accuracy against cost}
\label{sec:pareto}
% LogicTreeNet needs fewer gate operations than we do (61M against 3.8G for the ResNet-8, about 60$\times$) but stops at 86\% (\cref{tab:main}, \cref{fig:pareto} in the supplement).
% Among binary and ternary networks, ResoLUT is the cheapest at every accuracy from 86\% to 93\%, usually by 2--5$\times$.
% \todo{Replace the stored-bits comparison (Fig.~1) by gate operations.}

The price is gate operations, more than for small logic networks, but \method{} lies on a competitive accuracy--gate-operations Pareto front (\cref{fig:teaser}), closest to binary and hybrid networks.

\subsubsection{Throughput}
\label{sec:hardware}
% Among logic networks, LogicTreeNet-B has the fastest published FPGA implementation (80.2\%, 24\,ns per image), but LogicTreeNet-L and -G (85.0 and 86.3\%) were not implemented, so above 85\% on CIFAR-10 the comparison is with binary networks and LL-ViT (\cref{tab:fpga}).
% A streaming implementation of the ResoLUT-8 with one eighth of the trees (90.56\%, 223k LUT4 after pruning) on a Xilinx VU35P at 200\,MHz accepts an image every 5.3\,$\mu$s, \ie 189k images per second, set by the arrival of the $32{\times}32$ input.
% This is 63 to 1,100$\times$ faster than the binary networks above 85\% (67$\times$ for FracBNN at 89.1\%) and 170$\times$ faster than LL-ViT, which is 4.9 points more accurate; the devices differ.
% \todo{Latency and cycles of the 5.3\,$\mu$s design; resource use; cite H-WNN-L (88.2\%, 6.84\,ms).}
A streaming implementation of the ResoLUT-8 with one eighth of the trees (90.04\%$\pm$ 0.47\% on average over seeds, 223k LUT4 after pruning) on a Xilinx VU35P at 200\,MHz accepts an image every 5.3\,$\mu$s (189k images per second), set by the arrival of the $32{\times}32$ input (\cref{tab:fpga}).
This is 1.5$\times$ the throughput of Tridgell \etal, who are 0.9 points more accurate, and 66 to 1,180$\times$ that of the other networks, including the more accurate FireFly-T and LL-ViT; the devices differ.

\begin{table}[tb]
\centering
\footnotesize   
\setlength{\tabcolsep}{3pt}
\begin{tabular}{@{}lcrrr@{}}
\toprule
Method & Acc.\ (\%) & FPGA & Latency & Images/s \\
\midrule
ReBNet, 3 residuals~\cite{ghasemzadeh2018rebnet} & 86.98 & Zynq-7020 & -- & 2.0K \\
H-WNN-L~\cite{jadhao2025hybrid} & 88.23 & Zynq-7020 & 6.84\,ms & $\sim$0.8K \\
Zhao \etal~\cite{zhao2017accelerating} & 88.54 & Zynq-7020 & 5.94\,ms & 0.17K \\
FBNA CNV~\cite{guo2018fbna} & 88.61 & Zynq-7020 & -- & 0.52K \\
FracBNN~\cite{zhang2021fracbnn} & 89.10 & ZU3EG & -- & 2.8K \\
Tridgell \etal~\cite{tridgell2019unroll} & 90.9 & VU9P & 29\,$\mu$s & 122K \\
FireFly-T~\cite{li2025fireflyt} & 94.45 & ZU5EV & -- & 0.91K \\
LL-ViT~\cite{nag2025llvit} & 95.5 & VU9P & 5.33\,ms & 1.08K \\
\midrule
ResoLUT-8, 1/8 trees & 90.04 & VU35P & \textbf{5.3\,$\mu$s} & \textbf{189K} \\
\bottomrule
\end{tabular}
\caption{FPGA time per image, CIFAR-10 networks above 85\% test accuracy. Ours: VU35P, 200\,MHz.}
\label{tab:fpga}
\end{table}

% 
% \begin{table}[tb]
% \centering
% \footnotesize   
% \setlength{\tabcolsep}{3pt}
% \begin{tabular}{@{}lrrr@{}}
% \toprule
% Method & Acc.\ (\%) & FPGA t. & Images/s \\
% \midrule
% ReBNet, 3 residuals~\cite{ghasemzadeh2018rebnet} & 86.98 & 500\,$\mu$s & 2.0K \\
% H-WNN-L~\cite{jadhao2025hybrid} & 88.23 & 6.84\,ms & 2.87K \\
% Zhao et al.~\cite{zhao2017accelerating} & 88.54 & 5.94\,ms & 0.16K \\
% FBNA CNV~\cite{guo2018fbna} & 88.61 & 1.92\,ms & 0.52K \\
% FracBNN~\cite{zhang2021fracbnn} & 89.10 & 356\,$\mu$s & 2.8K \\
% Tridgell et al ~\cite{tridgell2019unroll} & 90.90 & 29\,$\mu$s & 122K \\
% FireFly-T~\cite{li2025fireflyt} & 94.45 & 1.1\,ms & 0.9K \\
% LL-ViT~\cite{nag2025llvit} & 95.50 & 0.92\,ms & 1.1K \\
% \midrule
% ResoLUT-8, 1/8 trees (ours) & 90.04 $\pm$ 0.47 & \textbf{5.3\,$\mu$s} & \textbf{189K} \\
% \bottomrule
% \end{tabular}
% \caption{FPGA time per image, CIFAR-10 networks above 85\% test accuracy. Ours: VU35P, 200\,MHz.}
% \label{tab:fpga}
% \end{table}

%% file: sec/5_discussion.tex
\section{Discussion and Limitations}
\label{sec:discussion}

\paragraph{What the results say.}
The additive shortcuts reduce conversion loss at greater depths, a hard forward pass removes the discretization gap, the wiring can be derived from the teacher, and a multi-bit residual stream carries the signal between layers.
The stream matters more the deeper the network, and it is nearly free: the residual path adds no LUTs, only wires and adder bits, and at 3 to 4 bits it costs nothing after brief fine-tuning. In contrast a network trained without residuals converts clearly worse from 20 layers on (3.9 against 3.1 points lost for the ResoLUT-20 and 5.0 against 3.2 for the ResoLUT-32 on CIFAR-100).
At 8 layers, however, a teacher without residuals converts almost as well, so the teacher's quality and the layer-wise procedure carry more of the result there; in the fully connected LILogic Net an additive residual stream also mitigates depth-related accuracy degradation but does not make deeper networks better.

\paragraph{Limitations.}
The teacher–student accuracy gap grows with the depth of the network (on CIFAR-10 from 1.2 points at 20 layers to 2.0 at 56, on ImageNet from 2.5 to 4.6 points from 20 to 56 layers), so a better optimization is needed to close this gap.
We did not test widely the optimizations connected with the residual stream and with the tree budget and shape, only for single examples; we leave this for future work.
Our FPGA results are limited to the 1/8-size ResoLUT-8; other, larger models on bigger FPGAs remain to be tried.

\section{Conclusion}
\label{sec:conclusion}
We have shown that LUT networks can be as deep as ResNets and close to them in accuracy if they are converted from one rather than trained from scratch.
ResoLUT reaches 95.66\% with twice the trees on CIFAR-10 and 78.87\% with the ResoLUT-56 on CIFAR-100, the best results of logic and lookup-table networks on both datasets, and it scales: accuracy keeps rising with depth on CIFAR-100 and ImageNet (53.69\% at 56 layers) and with the number of trees.
Its multi-bit residual stream matters more with depth, adds no LUTs and also removes the depth degradation of flat logic networks.
The results are promising: with one eighth of the trees and further optimizations the ResoLUT-8 already reaches 90.04\% on an FPGA at 189k images per second, and transferring these optimizations to the other models and budgets, promises further savings in LUTs with better accuracy.